\pdfoutput=1

\documentclass[11pt]{article}

\usepackage[final]{acl}

\usepackage{times}
\usepackage{latexsym}

\usepackage[T1]{fontenc}

\usepackage[utf8]{inputenc}

\usepackage{microtype}

\usepackage[utf8]{inputenc} 
\usepackage[T1]{fontenc}    
\usepackage{hyperref}       
\usepackage{url}            
\usepackage{booktabs}       
\usepackage{amsfonts}       
\usepackage{nicefrac}       
\usepackage{microtype}      
\usepackage{xcolor}         
\usepackage{enumitem}
\usepackage{float}

\usepackage{amsmath}
\usepackage{mathtools}
\usepackage{amsthm}
\usepackage{bm}
\usepackage{comment}
\usepackage{algorithm}
\usepackage{algpseudocode}
\usepackage{graphicx}
\usepackage{wrapfig}
\usepackage{blindtext}
\usepackage{lipsum}
\usepackage{setspace}
\usepackage{caption}
\usepackage{subcaption}
\usepackage{amssymb}
\usepackage{colonequals}
\usepackage{enumitem}
\usepackage{color, colortbl}
\usepackage{multirow}
\usepackage{tabularx, booktabs}
\usepackage{bbm}

\usepackage{hyperref}
\usepackage{url}
\usepackage{bm}
\usepackage{amsmath}
\usepackage{caption}
\usepackage{subcaption}
\usepackage{tikz}
\usepackage{arydshln}
\usepackage[symbol]{footmisc}
\usepackage{footnote}
\usepackage{multirow}
\usepackage[T1]{fontenc}    
\usepackage{graphicx}
\usepackage{wrapfig}
\usepackage{wrapfig,lipsum,booktabs}
\usepackage{amssymb}
\usepackage{array}
\usepackage{arydshln}
\usepackage{footnote}
\usepackage{placeins}
\usepackage[hyphenbreaks]{breakurl}
\usepackage{xurl}
\makesavenoteenv{tabular}
\makesavenoteenv{table}

\title{A Study on Knowledge Distillation from Weak Teacher\\ for Scaling Up Pre-trained Language Models}

\author{Hayeon Lee$^{1}$\thanks{\hspace{0.1in}Work done while interning at Meta AI.}\hspace{0.15in}Rui Hou$^{2}$\hspace{0.1in}Jongpil Kim$^{2}$\hspace{0.1in}
        Davis Liang$^{2}$\hspace{0.1in}Sung Ju Hwang$^{1}$\hspace{0.1in}Alexander Min$^{2}$\\
        KAIST$^{1}$\hspace{0.15in}Meta AI$^{2}$\\
        \normalsize \texttt{hayeon926@kaist.ac.kr rayhou@meta.com jpkim.ad@gmail.com}\\
        \normalsize \texttt{davis@abridge.com sjhwang82@kaist.ac.kr alexmin@meta.com} }

\begin{document}
\maketitle
\begin{abstract}
Distillation from Weak Teacher (DWT) is a method of transferring knowledge from a smaller, weaker teacher model to a larger student model to improve its performance. Previous studies have shown that DWT can be effective in the vision domain and natural language processing (NLP) pre-training stage. Specifically, DWT shows promise in practical scenarios, such as enhancing new generation or larger models using pre-trained yet older or smaller models and lacking a resource budget. However, the optimal conditions for using DWT have yet to be fully investigated in NLP pre-training. Therefore, this study examines three key factors to optimize DWT, distinct from those used in the vision domain or traditional knowledge distillation. These factors are:
(i) the impact of teacher model quality on DWT effectiveness, (ii) guidelines for adjusting the weighting value for DWT loss, and (iii) the impact of parameter remapping as a student model initialization technique for DWT.
\end{abstract}
\section{Introduction}
Recently, Distillation from Weak Teacher (DWT) \citep{yuan2020revisiting, qin-etal-2022-knowledge}, a reversed Knowledge Distillation (KD) technique, has gained attention from researchers. Unlike the traditional KD~\citep{distilbert2019, wang2020minilm, wang2020minilmv2, sun-etal-2019-patient, jiao-etal-2020-tinybert}, which compresses a pre-trained model by transferring its knowledge to a smaller model, DWT distills knowledge from a smaller (or weaker) pre-trained model to a larger model to improve its quality during training.

DWT is well-suited for practical real-world scenarios such as:
\begin{itemize}
\vspace{-0.05in}
\item Train a larger (scaled-up) model with an existing (smaller) pre-trained model to improve model quality using the same dataset.
\vspace{-0.05in}
\item Train a new, large-scale model with an old, smaller model to improve performance using the same dataset.
\vspace{-0.05in}
\item It is not feasible to use a large teacher model during KD training due to training resource constraints.
\vspace{-0.05in}
\end{itemize}
For the above cases, DWT can utilize the existing pre-trained models and improve the learning of new (larger) models.

Studies~\citep{yuan2020revisiting, qin-etal-2022-knowledge} have shown that DWT allows a larger student model to leverage the knowledge of a weaker, smaller pre-trained teacher model in both the computer vision and NLP pre-training stages. While previous research by~\citet{qin-etal-2022-knowledge} has demonstrated the potential of DWT in the NLP domain, it did not fully explore the key aspects of DWT such as the impact of teacher model quality and a student model initialization technique for DWT.

However, to truly unlock the potential of DWT for real-world applications, we need a deeper understanding of the key conditions and factors that contribute to its performance. For example, the effect of DWT might differ from traditional KD and potentially harm the student model, depending on the quality of its teacher.

Therefore, this work conducts in-depth studies and uncovers crucial insights to optimize DWT in the pre-training stage of NLP as follows:
\begin{figure*}[t]
    \small
    \centering
    \vspace{-0.05in}
    \includegraphics[width=0.98\textwidth]{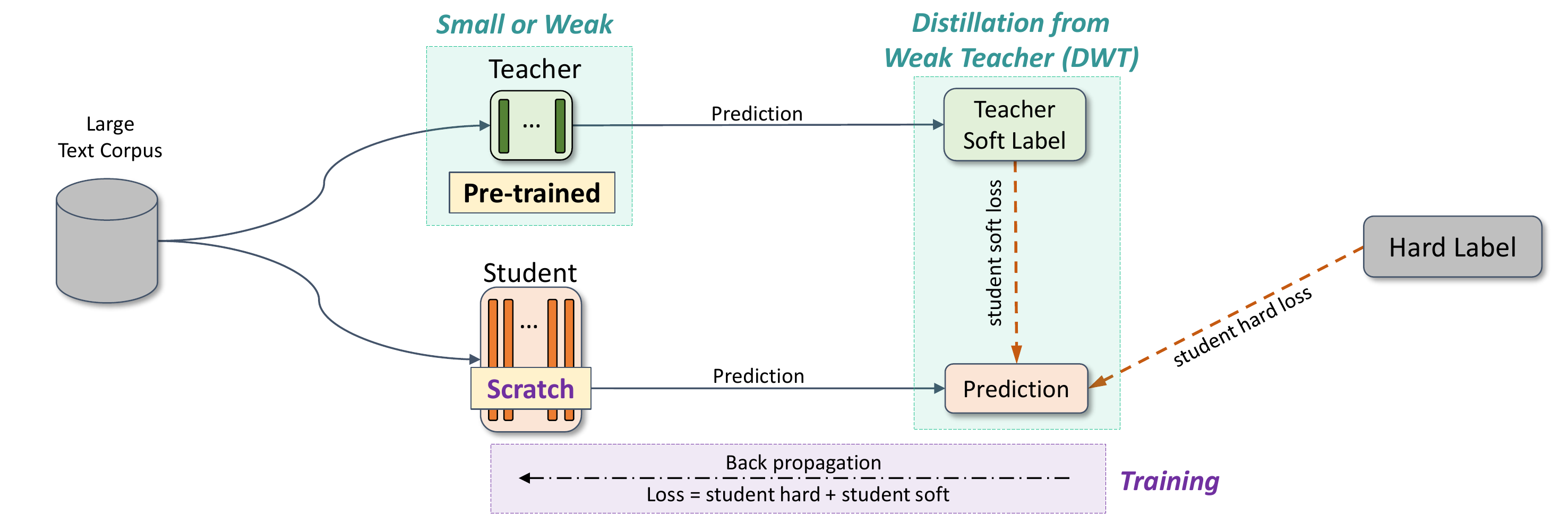} 
    \vspace{-0.1in}
    \caption{\small \textbf{Illustration of DWT Framework.} \small During the pre-training stage, the larger student model can learn from the knowledge and expertise of the \emph{small} or \emph{weak} teacher model,  enabling it to achieve better performance on various downstream tasks compared to training it standalone.
    }
    \label{fig:concept}
    \vspace{-0.15in}
\end{figure*}

\begin{itemize}
\vspace{-0.05in}
\item First, we investigate the effectiveness of DWT in relation to the quality of the teacher model. We find that an extremely weak teacher can negatively impact the student model's quality, which is different from the vision domain where even an extremely weak teacher still improves performance \citep{yuan2020revisiting}.
\vspace{-0.05in}
\item  Second, we examine the impact of distillation by adjusting the weighting value of the soft loss. We demonstrate that adjusting the weighting value for the DWT loss (soft loss) can improve training speed but may lead to suboptimal performance. To mitigate this issue, we recommend starting with a large weighting value and gradually decaying it during training.
\vspace{-0.05in}
\item Lastly, we study the effectiveness of Parameter Remapping (PR)~\citep{chen2015net2net, cai2018efficient, fang2020fast, leelightweight}, which is a popular student parameter initialization technique for conventional KD, as an initialization technique for DWT. We observe that PR leads to suboptimal solutions, contrary to its effectiveness in conventional KD scenarios. Random initialization is better than PR for DWT.
\end{itemize}
\vspace{-0.05in}
We believe that these observations provide useful guidelines to better utilize DWT techniques for real-world applications.

\section{Distillation from Weak Teacher}
In this section, we formulate the Distillation from Weak Teacher (DWT) strategy, which involves training the target (student) model using both the teacher's predictions (soft labels) and the ground truth (hard labels).
\paragraph{Task} 
Given a classification task with $c$ classes, for each training instance $x$ and its corresponding ground truth label $y$, the ground truth distribution over the labels is denoted as $q(c|x)$ (abbreviated as $q(c)$) where for each label $c$ in the set $\{ 1...C\}$, $q(y)=1$ and $q(c)=0$ for all $c$ not equal to $y$.

\paragraph{Model} The teacher model, with learnable parameters $\bm{\omega}$, and the student model, with learnable parameters $\bm{\theta}$, are utilized to predict the probability of each label $c$ for a given instance $x$. The probability predicted by the teacher model, denoted as $p^\tau_{\bm{\omega}}(c|x)$, and the probability predicted by the student model, denoted as $p^\tau_{\bm{\theta}}(c|x)$, are expressed as follows:
\begin{equation}
\begin{gathered}
p^{\tau}_{\bm{\omega}}(c | x) =softmax(\bm{z}^{\bm{\omega}}) = \frac{\exp(z_{c}^{\bm{\omega}}/\tau)}{\sum{^C_{i=1}}\exp(z^{\bm{\omega}}_i/\tau)} \nonumber
\end{gathered}
\end{equation}
\begin{equation}
\begin{gathered}
p^{\tau}_{\bm{\theta}}(c|x) =softmax(\bm{z}^{\bm{\theta}}) = \frac{\exp(z^{\bm{\theta}}_c/\tau)}{\sum{^C_{i=1}}\exp(z^{\bm{\theta}}_i/\tau)}  \nonumber
\end{gathered}
\end{equation}
where $\bm{z}^{\bm{\omega}} = \{z_i^{\bm{\omega}}\}_{i=1}^C$ is the output logit of the teacher model, $\bm{z}^{\bm{\theta}} = \{z_i^{\bm{\theta}}\}_{i=1}^C$ is the output logit of the student model, and $\tau$ is the temperature used to soften the probabilities  $p_{\bm{\omega}}(c)$ and $p_{\bm{\theta}}(c)$. 

\paragraph{Weak (Small) Teacher}
We assume that the parameter of the teacher model is pre-trained as $\bm{\omega}^*$. While conventional KD typically assumes that the size of the teacher model is larger than or equal to the size of the student model, i.e., $\lvert \bm{\omega}^* \rvert \geq \lvert \bm{\theta} \rvert$, DWT considers the case where the size of the teacher model is smaller than the size of the student model, i.e., $\lvert \bm{\omega}^* \rvert < \lvert \bm{\theta} \rvert$, or the quality of the pre-trained teacher model with parameters $\bm{\omega}^*$ is inferior to the quality of the pre-trained student model with parameters $\bm{\theta}^*$ obtained through stand-alone training.

\paragraph{Hard Loss}is the cross-entropy loss $H(q,p_{\bm{\theta}})$ between the ground truth $q$ and student's prediction $p_{\bm{\theta}}$, used to train the student model:
\begin{equation}
\begin{gathered}
H(q,p_{\bm{\theta}}) = - \sum_{c=1}^C q(c) \log (p_{\bm{\theta}}(c))
\label{eq:kd_loss}
\end{gathered}
\end{equation}
Following BERT~\citep{BERT}, $H(q,p_{\bm{\theta}})$ is the Masked Language Modeling loss (MLM loss).  
\paragraph{Soft Loss} is the Kullback-Leibler divergence (KL divergence) $S(p^\tau_{\bm{\omega}}, p^\tau_{\bm{\theta}})$ between the predictions of the student and the teacher models, and is given by:
\begin{equation}
\begin{gathered}
S(p^\tau_{\bm{\omega}}, p^\tau_{\bm{\theta}}) = \sum_{c=1}^C p^\tau_{\bm{\omega}} (c)  \cdot \log \frac{p^\tau_{\bm{\omega}}(c)}{p^\tau_{\bm{\theta}}(c)}, 
\label{eq:hard_loss}
\end{gathered}
\end{equation}
\paragraph{Final Objective} The objective function $\mathcal{L}(\bm{\theta})$ aims to train the student model by minimizing a weighted sum of the hard loss and the soft loss:
\begin{equation}
\begin{gathered}
\mathcal{L}(\bm{\theta}) = \alpha_{h} \cdot H(q, p_{\bm{\theta}}) +\alpha_{s} \cdot S(p^\tau_{\bm{\omega}}, p^\tau_{\bm{\theta}})
\label{eq:kd_loss}
\end{gathered}
\end{equation}
where the weighting hyperparameters for the hard loss and the soft loss are denoted by $\alpha_{h}$ and $\alpha_{s}$, respectively.
\begin{figure}[t]
    \small
    \centering
    \vspace{-0.05in}
    \includegraphics[width=\columnwidth]{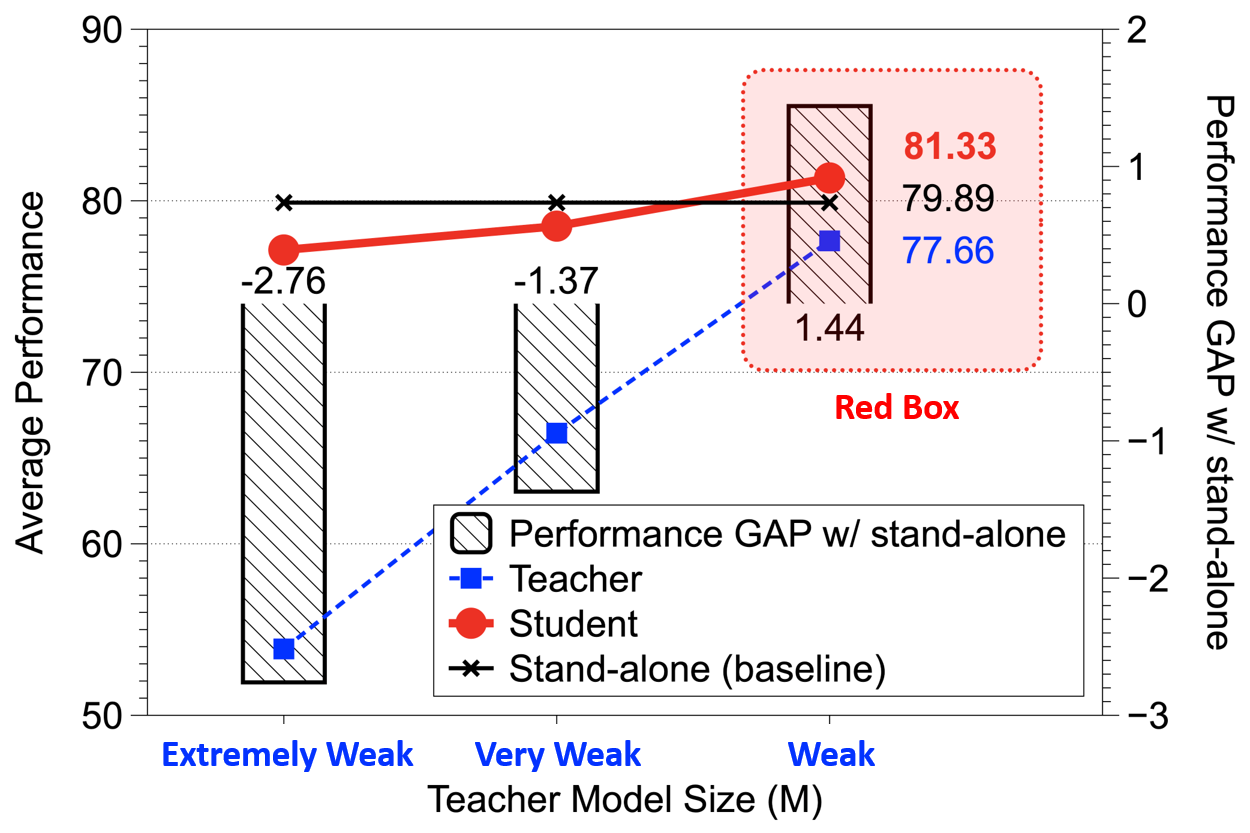} 
    \vspace{-0.1in}
    \caption{\small \textbf{Impact of Teacher Quality} \small [Red Box] The Weak teacher model significantly improves the performance of the student model by 1.44, increasing from 79.89 to 81.33. However, distillation from Very Weak or Extremely Weak teachers has a negative impact on the performance of the student.}
    \label{fig:main}
    \vspace{-0.15in}
\end{figure}

\section{Experiment}
\begin{figure*}[t]
    \small
    \centering
    \vspace{-0.05in}
    \includegraphics[width=0.495\columnwidth]{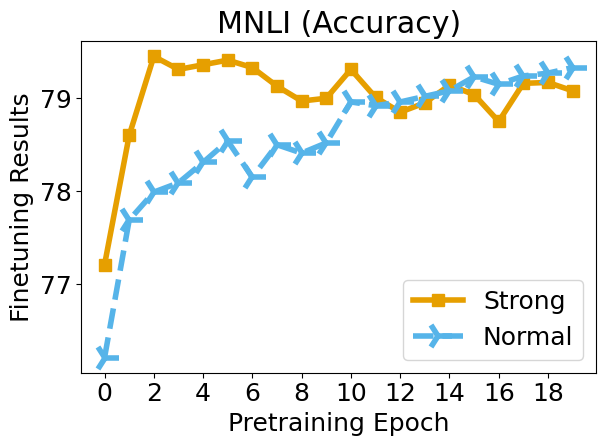} 
    \includegraphics[width=0.495\columnwidth]{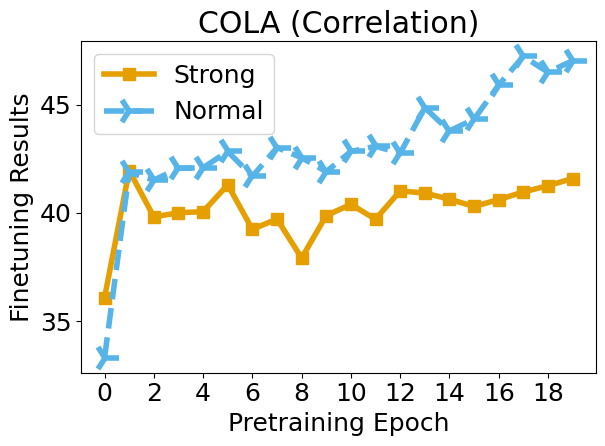} 
    \includegraphics[width=0.495\columnwidth]{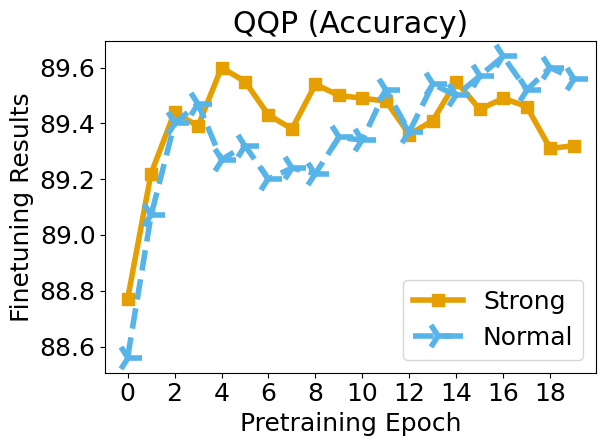} 
    \includegraphics[width=0.495\columnwidth]{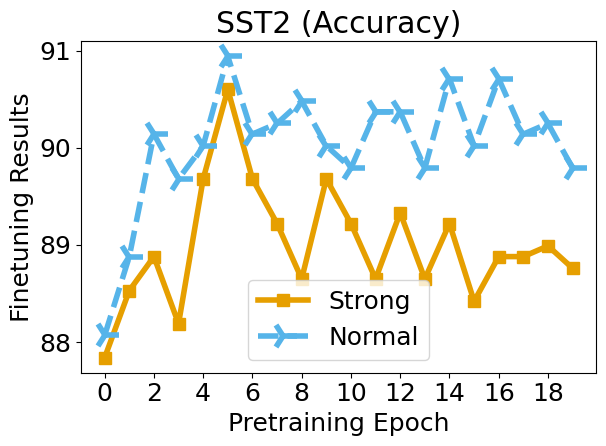} 
    \vspace{-0.1in}
    \caption{\small \textbf{Adjusting Soft Loss Weight} \small 
    Unlike conventional KD, where using large weights for the soft loss improves training convergence speed and performance, DWT requires careful tuning of the loss weight. Using a large weight leads to faster convergence, but a small weight leads to better fine-tuning performance.}
    \label{fig:weight}
\end{figure*}
\begin{figure*}[t]
    \small
    \centering
    \includegraphics[width=0.495\columnwidth]{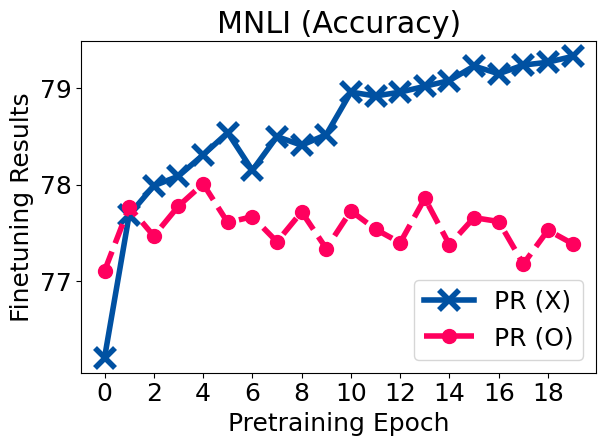} 
    \includegraphics[width=0.495\columnwidth]{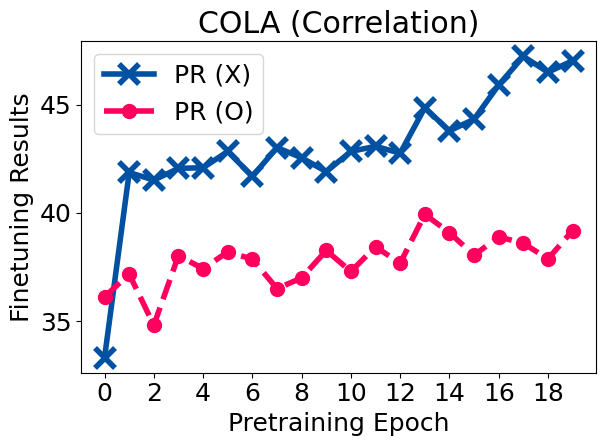} 
    \includegraphics[width=0.495\columnwidth]{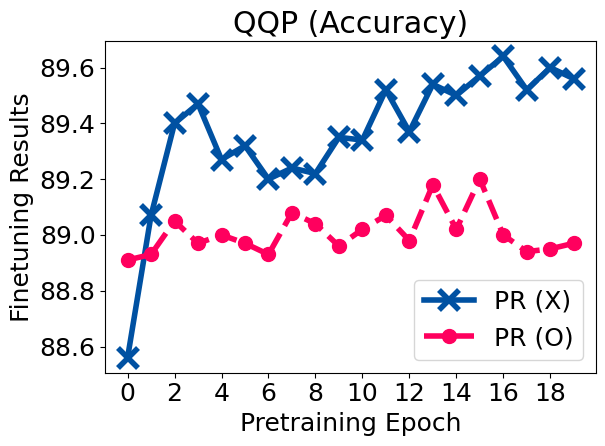} 
    \includegraphics[width=0.495\columnwidth]{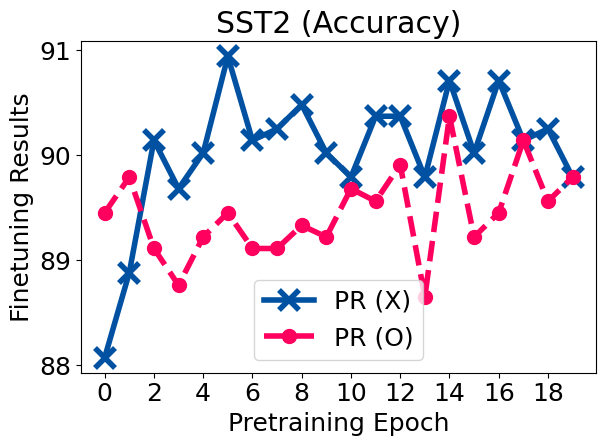} 
    \vspace{-0.1in}
    \caption{\small \textbf{Impact of Parameter Remapping} \small 
    Unlike in conventional KD training, where parameter remapping (PR) (\texttt{PR(O)}) is effective, we found that PR hinders DWT training, leading to local optima. Even with continued pre-training, the fine-tuning performance does not improve. Therefore, random initialization (\texttt{PR(X)}) appears to be more beneficial for DWT.}
    \label{fig:pr}
    \vspace{-0.15in}
\end{figure*}

We conducted a study to analyze the efficacy of the DWT method and present key observations for optimizing its impact in three core elements: (i) the quality of the teacher model, (ii) the degree of soft knowledge transfer, and (iii) the initialization type (parameter remapping) of the student model.
\paragraph{Training setting} we use a default loss weight ratio of $\alpha_h:\alpha_s$ = 1:1 for the hard loss and soft loss during distillation. The learning rate is set to $5e-4$, and the models are trained for 20 epochs with the application of quantization, linear warm-up (5\%), the Adam optimizer~\citep{kingma2014adam}, 16 batch sizes per GPU, and 8 A100 GPUs per run. In the pre-training stage, we utilize a reduced dataset of 30 million sentences generated by uniformly selecting one sentence out of every four sentences from the original dataset, which consists of a combination of BookCorpus~\citep{zhu2015aligning} and Wikipedia~\citep{wikidump}. The performance of the distilled models is evaluated on the dev sets of the GLUE benchmark~\citep{wang2018glue}, comprising nine sentence-level classification tasks (Please see the supplementary file for more details.).

\subsection{Impact of Teacher Model Quality}
\label{exp:main}
In Figure~\ref{fig:main}, we examine the performance of distilled student models based on the quality of the teacher model. We conduct a distillation from a teacher model during the pre-training stage and fine-tune the distilled student models on the dev sets of the GLUE benchmark. We report the average performance and the performance gap between the distilled student and a student trained standalone. We categorize the weak teacher quality into three levels compared to the standalone student model, which has a model size of 67M and achieves an average performance of 79.89 on the GLUE benchmark dev sets. 

\textbf{1) Weak}: 0.78$\times$ smaller size, -2.23 lower performance

\textbf{2) Very Weak}: 0.57$\times$ smaller size, -13.44 lower performance

\textbf{3) Extremely Weak}: 0.46$\times$ smaller size, -26.02 lower performance.

While distillation from weak teachers, even extremely weak ones, consistently improves the performance of the student model in the vision field due to the regularization effect~\citep{yuan2020revisiting}, we found that in language model pre-training, the effectiveness of DWT on the student model heavily depends on the quality of the teacher model. The student model (the red mark) clearly benefits from the \textbf{Weak} teacher model (score is 77.66), represented by the blue mark in the red box, as it shows an improvement of 1.44 points, from 79.89 to 81.33. However, when the performance gap between the teacher and student is too large, such as in the cases of \textbf{Very Weak} and \textbf{Extremely Weak} teachers, distillation from these teachers may negatively impact the student's performance by -1.37 and -2.76, respectively. Our observations provide valuable guidance for researchers aiming to utilize existing pre-trained models in training new models.

\subsection{The Impact of Soft Loss} 
\label{exp:loss_weight}

In Figure~\ref{fig:weight}, we investigate the impact of the soft loss in DWT during the pre-training stage by adjusting the weights in the following two versions: \textbf{(1) Strong}: We fix the weight for the hard loss at 1 and multiply the weight for the soft loss by 4 to increase the intensity of distillation. \textbf{(2) Normal}: The ratio between the hard loss and soft loss is equal, with the soft loss weight set to 1. Finally, we fine-tune the models pre-trained with different soft loss weights on the GLUE benchmark tasks.

Conventional KD has shown that using large weights for the soft loss can improve both training convergence and model performance~\citep{distilbert2019}. However, we reveal that DWT requires careful tuning of the soft weights. Our observations show that using a large weight for the soft loss (\textbf{Strong}) leads to faster convergence in most downstream tasks (e.g., MNLI~\cite{williams-etal-2018-broad}, COLA~\cite{warstadt-etal-2019-neural}, QQP~\cite{WinNT}, SST2~\cite{SST-5dataset2013}) compared to using a small weight (\textbf{Normal}). However, as training continues, using a small weight for the soft loss (\textbf{Normal}) leads to better fine-tuning performance than using a large weight (\textbf{Strong}). Therefore, we believe that gradually decreasing the soft loss weights (e.g., from 4 to 1) during training would benefit both convergence and performance.

\subsection{Impact of Parameter Remapping} 
\label{exp:pr}
Parameter remapping (PR)~\citep{chen2015net2net, cai2018efficient, fang2020fast, fna++} is a popular technique used in conventional KD methods. It involves copying the parameters of a pre-trained teacher model and initializing the student model with these parameters before starting KD training (See the supplementary file for more details.). PR can accelerate convergence speed and improve the final performance of the distilled model. For example, DistilBERT~\citep{distilbert2019} uniformly samples six layers out of twelve from the BERT model (teacher) and initializes the corresponding layers in DistilBERT (student) with the copied parameters. 

In Figure~\ref{fig:pr}, we investigate the effectiveness of PR for knowledge transfer from a smaller model to a larger model. Before DWT training, we copy parameters from the first four layers of the teacher model and paste them into the corresponding layers of the student model. Following the approach of~\citet{fang2020fast, fna++}, we also use the parameters of the last layer in the teacher model for the remaining fifth and sixth layers of the student model.

We initialize student models with PR (\texttt{PR(O)}) or randomly (\texttt{PR(X)}), train them with distillation on a large text corpus, and fine-tune the distilled student models on various downstream tasks. Experimental results show that, unlike in conventional KD training, PR (\texttt{PR(O)}) hinders DWT training, leading to local optima. With PR, the performance of the fine-tuned models does not improve even with continued pre-training. Therefore, random initialization (\texttt{PR(X)}) is more beneficial for DWT.
\section{Conclusion}
Distillation from Weak Teacher (DWT) is a technique that improves the performance of a larger student model by transferring knowledge from a weaker, smaller teacher model. Despite the potential of DWT, the optimal conditions to use DWT have yet to be fully investigated in NLP pre-training. This study investigated three crucial factors for optimizing DWT in NLP pre-training, which differ from those in vision or traditional KD. These factors include the impact of teacher model quality, the use of parameter remapping as an initialization technique for DWT, and guidelines for adjusting the weighting value of the DWT loss.

\section*{Limitations}
In this section, we faithfully discuss the current limitations and potential avenues for future research. First of all, in the analysis, we observed that giving heavy weight to the soft loss at initial training epochs improves the convergence speed. Yet, continuing training with such heavy weight to the soft loss could hinder the further performance improvement of the student. Therefore, adjusting soft loss weights depending on the training phase from a larger value to a small value (e.g., using the time function) would be helpful for both convergence speed and improving the model’s quality.

Secondly, it has been demonstrated in the visual recognition domain that adjusting the temperature of distillation loss for poorly performed teachers can improve the student model quality due to the regularization effect.
Following them, increasing the temperature to smooth the soft labels from poorly performed teachers, such as 1-layer or 2-layer teachers, would help improve the quality of distillation via the regularization effect.

\section*{Ethics Statement}
Our Distillation from Weak Teacher (DWT) framework facilitates enhancing larger student models through knowledge transfer from smaller, weaker teacher models. However, our research findings indicate that the effectiveness of the teacher model, particularly when it is extremely weak, can have a negative impact on the quality of the student model. Consequently, the utilization of our DWT framework should be approached with caution, particularly in high-risk domains like biomedicine. Evaluating performance prior to making critical decisions may be necessary.
\bibliography{anthology,custom}

\begin{thebibliography}{27}
\expandafter\ifx\csname natexlab\endcsname\relax\def\natexlab#1{#1}\fi

\bibitem[{Cai et~al.(2018)Cai, Chen, Zhang, Yu, and Wang}]{cai2018efficient}
Han Cai, Tianyao Chen, Weinan Zhang, Yong Yu, and Jun Wang. 2018.
\newblock Efficient architecture search by network transformation.
\newblock In \emph{Proceedings of the AAAI Conference on Artificial
  Intelligence}.

\bibitem[{Cer et~al.(2017)Cer, Diab, Agirre, Lopez-Gazpio, and
  Specia}]{cer2017semeval}
Daniel Cer, Mona Diab, Eneko Agirre, Inigo Lopez-Gazpio, and Lucia Specia.
  2017.
\newblock Semeval-2017 task 1: Semantic textual similarity-multilingual and
  cross-lingual focused evaluation.
\newblock \emph{arXiv preprint arXiv:1708.00055}.

\bibitem[{Chen et~al.(2015)Chen, Goodfellow, and Shlens}]{chen2015net2net}
Tianqi Chen, Ian Goodfellow, and Jonathon Shlens. 2015.
\newblock Net2net: Accelerating learning via knowledge transfer.
\newblock \emph{arXiv preprint arXiv:1511.05641}.

\bibitem[{Dagan et~al.(2005)Dagan, Glickman, and Magnini}]{dagan2005pascal}
Ido Dagan, Oren Glickman, and Bernardo Magnini. 2005.
\newblock The pascal recognising textual entailment challenge.
\newblock In \emph{Machine Learning Challenges Workshop}, pages 177--190.
  Springer.

\bibitem[{Devlin et~al.(2019)Devlin, Chang, Lee, and Toutanova}]{BERT}
Jacob Devlin, Ming{-}Wei Chang, Kenton Lee, and Kristina Toutanova. 2019.
\newblock {BERT:} pre-training of deep bidirectional transformers for language
  understanding.
\newblock In \emph{Proceedings of the Conference of the North American Chapter
  of the Association for Computational Linguistics: Human Language
  Technologies}.

\bibitem[{Dolan and Brockett(2005)}]{dolan2005automatically}
William~B Dolan and Chris Brockett. 2005.
\newblock Automatically constructing a corpus of sentential paraphrases.
\newblock In \emph{Proceedings of the Third International Workshop on
  Paraphrasing (IWP2005)}.

\bibitem[{Fang et~al.(2020{\natexlab{a}})Fang, Sun, Peng, Zhang, Li, Liu, and
  Wang}]{fang2020fast}
Jiemin Fang, Yuzhu Sun, Kangjian Peng, Qian Zhang, Yuan Li, Wenyu Liu, and
  Xinggang Wang. 2020{\natexlab{a}}.
\newblock Fast neural network adaptation via parameter remapping and
  architecture search.
\newblock In \emph{International Conference on Learning Representations}.

\bibitem[{Fang et~al.(2020{\natexlab{b}})Fang, Sun, Zhang, Peng, Li, Liu, and
  Wang}]{fna++}
Jiemin Fang, Yuzhu Sun, Qian Zhang, Kangjian Peng, Yuan Li, Wenyu Liu, and
  Xinggang Wang. 2020{\natexlab{b}}.
\newblock \href {https://doi.org/10.1109/TPAMI.2020.3044416} {Fna++: Fast
  network adaptation via parameter remapping and architecture search}.
\newblock \emph{IEEE Transactions on Pattern Analysis and Machine
  Intelligence}.

\bibitem[{Foundation()}]{wikidump}
Wikimedia Foundation.
\newblock \href {https://dumps.wikimedia.org} {Wikimedia downloads}.

\bibitem[{Iyer et~al.(2017)Iyer, Dandekar, and Csernai}]{WinNT}
Shankar Iyer, Nikhil Dandekar, and Kornel Csernai. 2017.
\newblock \href
  {https://data.quora.com/First-Quora-Dataset-Release-Question-Pairs} {First
  quora dataset release: Question pairs}.

\bibitem[{Jiao et~al.(2020)Jiao, Yin, Shang, Jiang, Chen, Li, Wang, and
  Liu}]{jiao-etal-2020-tinybert}
Xiaoqi Jiao, Yichun Yin, Lifeng Shang, Xin Jiang, Xiao Chen, Linlin Li, Fang
  Wang, and Qun Liu. 2020.
\newblock {T}iny{BERT}: Distilling {BERT} for natural language understanding.
\newblock In \emph{Findings of Proceedings of the Conference on Empirical
  Methods in Natural Language Processing}.

\bibitem[{Kingma and Ba(2014)}]{kingma2014adam}
Diederik~P Kingma and Jimmy Ba. 2014.
\newblock Adam: A method for stochastic optimization.
\newblock \emph{arXiv preprint arXiv:1412.6980}.

\bibitem[{Lee et~al.(2022)Lee, An, Kim, and Hwang}]{leelightweight}
Hayeon Lee, Sohyun An, Minseon Kim, and Sung~Ju Hwang. 2022.
\newblock Lightweight neural architecture search with parameter remapping and
  knowledge distillation.
\newblock In \emph{First Conference on Automated Machine Learning
  (Late-Breaking Workshop)}.

\bibitem[{Levesque et~al.(2012)Levesque, Davis, and
  Morgenstern}]{levesque2012winograd}
Hector Levesque, Ernest Davis, and Leora Morgenstern. 2012.
\newblock The winograd schema challenge.
\newblock In \emph{Thirteenth International Conference on the Principles of
  Knowledge Representation and Reasoning}.

\bibitem[{Qin et~al.(2022)Qin, Lin, Yi, Zhang, Han, Zhang, Su, Liu, Li, Sun,
  and Zhou}]{qin-etal-2022-knowledge}
Yujia Qin, Yankai Lin, Jing Yi, Jiajie Zhang, Xu~Han, Zhengyan Zhang, Yusheng
  Su, Zhiyuan Liu, Peng Li, Maosong Sun, and Jie Zhou. 2022.
\newblock Knowledge inheritance for pre-trained language models.
\newblock In \emph{Proceedings of the Conference of the North American Chapter
  of the Association for Computational Linguistics: Human Language
  Technologies}.

\bibitem[{Rajpurkar et~al.(2016)Rajpurkar, Zhang, Lopyrev, and
  Liang}]{rajpurkar2016squad}
Pranav Rajpurkar, Jian Zhang, Konstantin Lopyrev, and Percy Liang. 2016.
\newblock Squad: 100,000+ questions for machine comprehension of text.
\newblock \emph{arXiv preprint arXiv:1606.05250}.

\bibitem[{Sanh et~al.(2019)Sanh, Debut, Chaumond, and Wolf}]{distilbert2019}
Victor Sanh, Lysandre Debut, Julien Chaumond, and Thomas Wolf. 2019.
\newblock Distilbert, a distilled version of bert: smaller, faster, cheaper and
  lighter.
\newblock \emph{arXiv preprint arXiv:1910.01108}.

\bibitem[{Socher et~al.(2013)Socher, Perelygin, Wu, Chuang, Manning, Ng, and
  Potts}]{SST-5dataset2013}
Richard Socher, Alex Perelygin, Jean Wu, Jason Chuang, Christopher~D Manning,
  Andrew Ng, and Christopher Potts. 2013.
\newblock Recursive deep models for semantic compositionality over a sentiment
  treebank.
\newblock In \emph{Proceedings of the Conference on Empirical Methods in
  Natural Language Processing}.

\bibitem[{Sun et~al.(2019)Sun, Cheng, Gan, and Liu}]{sun-etal-2019-patient}
Siqi Sun, Yu~Cheng, Zhe Gan, and Jingjing Liu. 2019.
\newblock Patient knowledge distillation for {BERT} model compression.
\newblock In \emph{Proceedings of the Conference on Empirical Methods in
  Natural Language Processing and the International Joint Conference on Natural
  Language Processing}.

\bibitem[{Wang et~al.(2019)Wang, Singh, Michael, Hill, Levy, and
  Bowman}]{wang2018glue}
Alex Wang, Amanpreet Singh, Julian Michael, Felix Hill, Omer Levy, and
  Samuel~R. Bowman. 2019.
\newblock \href {https://openreview.net/forum?id=rJ4km2R5t7} {{GLUE}: A
  multi-task benchmark and analysis platform for natural language
  understanding}.
\newblock In \emph{International Conference on Learning Representations}.

\bibitem[{Wang et~al.(2020{\natexlab{a}})Wang, Bao, Huang, Dong, and
  Wei}]{wang2020minilmv2}
Wenhui Wang, Hangbo Bao, Shaohan Huang, Li~Dong, and Furu Wei.
  2020{\natexlab{a}}.
\newblock Minilmv2: Multi-head self-attention relation distillation for
  compressing pretrained transformers.
\newblock \emph{arXiv preprint arXiv:2012.15828}.

\bibitem[{Wang et~al.(2020{\natexlab{b}})Wang, Wei, Dong, Bao, Yang, and
  Zhou}]{wang2020minilm}
Wenhui Wang, Furu Wei, Li~Dong, Hangbo Bao, Nan Yang, and Ming Zhou.
  2020{\natexlab{b}}.
\newblock Minilm: Deep self-attention distillation for task-agnostic
  compression of pre-trained transformers.
\newblock \emph{Advances in Neural Information Processing Systems}.

\bibitem[{Warstadt et~al.(2019)Warstadt, Singh, and
  Bowman}]{warstadt-etal-2019-neural}
Alex Warstadt, Amanpreet Singh, and Samuel~R. Bowman. 2019.
\newblock Neural network acceptability judgments.
\newblock \emph{Transactions of the Association for Computational Linguistics}.

\bibitem[{Williams et~al.(2018)Williams, Nangia, and
  Bowman}]{williams-etal-2018-broad}
Adina Williams, Nikita Nangia, and Samuel~R Bowman. 2018.
\newblock A broad-coverage challenge corpus for sentence understanding through
  inference.
\newblock In \emph{Association for Computational Linguistics}.

\bibitem[{Wolf et~al.(2020)Wolf, Debut, Sanh, Chaumond, Delangue, Moi, Cistac,
  Rault, Louf, Funtowicz, Davison, Shleifer, von Platen, Ma, Jernite, Plu, Xu,
  Le~Scao, Gugger, Drame, Lhoest, and Rush}]{wolf-etal-2020-transformers}
Thomas Wolf, Lysandre Debut, Victor Sanh, Julien Chaumond, Clement Delangue,
  Anthony Moi, Pierric Cistac, Tim Rault, Remi Louf, Morgan Funtowicz, Joe
  Davison, Sam Shleifer, Patrick von Platen, Clara Ma, Yacine Jernite, Julien
  Plu, Canwen Xu, Teven Le~Scao, Sylvain Gugger, Mariama Drame, Quentin Lhoest,
  and Alexander Rush. 2020.
\newblock Transformers: State-of-the-art natural language processing.
\newblock In \emph{Proceedings of the Conference on Empirical Methods in
  Natural Language Processing: System Demonstrations}.

\bibitem[{Yuan et~al.(2020)Yuan, Tay, Li, Wang, and Feng}]{yuan2020revisiting}
Li~Yuan, Francis~EH Tay, Guilin Li, Tao Wang, and Jiashi Feng. 2020.
\newblock Revisiting knowledge distillation via label smoothing regularization.
\newblock In \emph{Proceedings of the IEEE/CVF Conference on Computer Vision
  and Pattern Recognition}.

\bibitem[{Zhu et~al.(2015)Zhu, Kiros, Zemel, Salakhutdinov, Urtasun, Torralba,
  and Fidler}]{zhu2015aligning}
Yukun Zhu, Ryan Kiros, Rich Zemel, Ruslan Salakhutdinov, Raquel Urtasun,
  Antonio Torralba, and Sanja Fidler. 2015.
\newblock Aligning books and movies: Towards story-like visual explanations by
  watching movies and reading books.
\newblock In \emph{Proceedings of the IEEE international conference on computer
  vision}.

\end{thebibliography}

\appendix
\newpage

\section{Additional Experimental Setups}
\label{appendix:setups}

\subsection{Datasets}
We evaluate our proposed model's natural language understanding capabilities using the General Language Understanding Evaluation (GLUE) benchmark~\cite{wang2018glue}. GLUE consists of 9 sentence- or sentence-pair language understanding tasks, selected to cover a range of dataset sizes, text genres, and difficulty levels. We validate our pre-trained models on 7 downstream tasks: MNLI~\cite{williams-etal-2018-broad}, QQP~\cite{WinNT}, QNLI~\cite{rajpurkar2016squad}, SST-2~\cite{SST-5dataset2013}, CoLA~\cite{warstadt-etal-2019-neural}, STSB~\cite{cer2017semeval}, and MRPC~\cite{dolan2005automatically}. We exclude the smallest datasets, RTE~\cite{dagan2005pascal} and WNLI~\cite{levesque2012winograd}, due to their unstable training trends caused by the lack of training samples. When evaluating a model's performance on GLUE, the tasks are treated as individual binary classification problems, and an overall performance score is computed. The models are trained on large-scale datasets and evaluated based on accuracy or task-specific metrics. Detailed descriptions of each dataset that we used are as follows:

\textbf{MNLI (Multi-Genre Natural Language Inference)} task: Models make predictions about the logical relationship (entailment, contradiction, or neutral) between a premise and a hypothesis.

\textbf{CoLA (Corpus of Linguistic Acceptability)} task: The task focuses on judging grammaticality by classifying whether a given sentence is linguistically acceptable or not.

\textbf{QQP (Quora Question Pairs)} task: The objective is to determine the semantic equivalence of a pair of questions.

\textbf{SST-2 (Stanford Sentiment Treebank)} task: The goal is sentence-level sentiment classification, where models predict the sentiment polarity (positive or negative) of a given sentence.

\textbf{MRPC (Microsoft Research Paraphrase Corpus)} task: Models ascertain whether a pair of sentences are semantically equivalent or not.

\textbf{STS-B (Semantic Textual Similarity Benchmark)} task: The task aims to measure the semantic similarity between pairs of sentences using a continuous similarity score.

\textbf{QNLI (Question-answering Natural Language Inference)} task: The objective is to answer questions based on a given context paragraph, determining whether a question can be answered using the provided context.








\subsection{Experimental Details}
In Figure 2, we validate our pre-trained models on 7 downstream tasks: MNLI, QQP, QNLI, SST-2, CoLA, STSB, and MRPC of the GLUE benchmark. We fine-tune the models on each downstream task and calculate the average performance across all 7 tasks. In Figures 3 and 4, we report the performance of the models on individual tasks such as MNLI, CoLA, QQP, and SST-2 tasks. 

\paragraph{Fine-tuning Setups}
For fine-tuning models on text-classification tasks using the PyTorch framework, we used the code provided by Hugging Face~\cite{wolf-etal-2020-transformers}. The code can be found at \href{https://github.com/huggingface/transformers/tree/main/examples/pytorch/text-classification}{https://github.com/huggingface/transformers}. Our fine-tuning process followed the guidelines provided in the code, which included a learning rate of 2e-5, 3 epochs for most tasks in the GLUE benchmark~\cite{wang2018glue}, except for MRPC~\cite{dolan2005automatically} task where we used 5 fine-tuning epochs. We used a batch size of 32 and a maximum sequence length of 128 for all tasks. 

\subsection{Details of Parameter Remapping}
Following~\cite{fang2020fast}, we employed layer-wise parameter remapping to adjust the parameters of a small source model (teacher) to match the structure and size of a larger target model (student). The goal is to efficiently transfer the knowledge from the pre-trained source model to the larger target model. The process involves two key steps:

\textbf{Expansion}: In this step, the teacher model is expanded to match the size and structure of the larger student model. The expansion process includes duplicating the last layer and its parameters of the teacher model as needed to ensure compatibility with the student model. This expansion ensures that the teacher model has the same layer configuration as the student model, providing a foundation for subsequent parameter remapping.

\textbf{Remapping}:
Once the teacher model has been expanded, the next step is to remap the parameters between the expanded teacher model and the student model. The remapping is performed by mapping the parameters of the teacher model to the corresponding layer position of the student model.

\end{document}